\documentclass{bmvc2k}

\title{Orthogonal Polynomial Approximation for\\
Matrix Log Normalization in Global \\ Covariance Pooling}

\addauthor{Md Rifat Ur Rahman}{1806033@eee.buet.ac.bd}{1,2}
\addauthor{Md Raihan Khan}{khan2403558@stud.kuet.ac.bd}{1,3}
\addauthor{Md Sakib Hossain Shovon}{sakib@kaist.ac.kr}{4}
\addauthor{Pietro Li\`o}{pl219@cam.ac.uk}{5}
\addauthor{Mohammad Ali Moni}{mmoni@csu.edu.au}{1,6}
\addinstitution{
 NeuronTreeAI
}
\addinstitution{
 Bangladesh University of Engineering \\ and Technology
}
\addinstitution{
 Khulna University of Engineering and \\ Technology
}
\addinstitution{
 KAIST
}
\addinstitution{
 University of Cambridge
}
\addinstitution{
 Charles Sturt University
}

\runninghead{Rahman \etal}{Polynomial Matrix Log Normalization}

\def\etal{\emph{et al}\bmvaOneDot}

\usepackage{booktabs}
\usepackage{multirow}
\usepackage{makecell}
\usepackage{amsmath}
\usepackage{amssymb}
\usepackage{placeins}
\usepackage{url}
\usepackage{float}                 
\usepackage{algorithm}
\usepackage[noend]{algpseudocode}

\newcommand{\trace}{\operatorname{tr}}
\newcommand{\diag}{\operatorname{diag}}
\newcommand{\sym}{\operatorname{sym}}

\begin{document}

\maketitle

\begin{abstract}
Global Covariance Pooling (GCP) improves deep networks by capturing
second-order feature statistics, and is especially effective for
fine-grained recognition. Because covariance matrices live on the
Symmetric Positive Definite (SPD) manifold, a normalization step is
required before the Euclidean classifier. The faithful choice is the
matrix logarithm (MLN-COV), which maps the SPD manifold to its tangent
space; in practice it was abandoned in favour of the matrix square root
because its eigendecomposition-based gradient is numerically unstable. We
show that this instability is an artifact of computing the logarithm
\emph{spectrally}, not of the logarithm itself. Approximating the
logarithm with finite polynomials in the covariance matrix removes the
eigendecomposition from both passes: every operation becomes a General
Matrix Multiplication (GEMM), the gradient stays bounded on the spectral
support of the pre-normalized covariance, and the unstable
$1/(\lambda_i-\lambda_j)$ term never appears. The key ingredient is a
\emph{mean-eigenvalue} pre-normalization that centres the spectrum near
$1$, away from the singularity of $\log$, with a scalar post-compensation
that returns the singular part of $\log(A)$ in closed form. Our
recommended normalizer is a degree-8 Chebyshev expansion evaluated by a
three-term matrix recurrence, with a matching reverse recurrence for the
backward pass; Legendre, Laguerre, Taylor and Pad\'e expansions are
studied as controls that isolate the roles of the basis and of the target
function. On three fine-grained benchmarks and ImageNet-1k the
decomposition-free logarithm is both faster and more accurate than the
spectral logarithm and than the square-root approximations it replaces,
and at matched basis and degree the log target beats the square-root
target, confirming that the gain comes from the faithful Riemannian map
rather than from a better polynomial family. Code:
\url{https://github.com/RifatRahmanRimon/GCP_Orthogonal_Polynomial}.
\end{abstract}

\section{Introduction}
\label{sec:intro}

Global Covariance Pooling (GCP) was introduced as a second-order
alternative to global average pooling~\cite{DeepO2P}. Proposed first for
fine-grained visual classification
(FGVC)~\cite{GCP-FGVC,GCP-FGVC2,MOMN_FGVC_2020,Eigen_FGVC_2022}, it has
since been used in facial expression recognition~\cite{GCPFace},
histopathology~\cite{GCPbreast}, hyperspectral
imaging~\cite{GCPHyperSpectral}, SAR classification~\cite{GCPSAR,GCPSAR2}
and object detection~\cite{GCPObjectDetection}: by modelling channel-wise
correlations, covariance representations are more discriminative than
first-order statistics. Given final-layer activations
$X\in\mathbb{R}^{d\times N}$, GCP computes the covariance
$A = X\bar IX^\top$ with
$\bar I=\tfrac{1}{N}(I-\tfrac{1}{N}\mathbf{1}\mathbf{1}^\top)$ centring
the features. The result $A$ lies on the Symmetric Positive Definite
(SPD) manifold, whereas the classifier operates in Euclidean space, so a
normalization step is required; these normalizers can be read as implicit
Riemannian classifiers~\cite{Riemann}.

\paragraph{The faithful map and why it was abandoned.}
The faithful normalization is the matrix logarithm (MLN-COV),
\begin{equation}
\hat A_{\mathrm{MLN}} = \log(A) = U\log(\Lambda)U^\top,
\qquad A=U\Lambda U^\top,
\label{eq:mln}
\end{equation}
which maps the SPD manifold onto its tangent space at the identity. It is
rarely used, because Eq.~\eqref{eq:mln} requires an eigendecomposition
(EIG/SVD) and its gradient is fragile. MPN-COV~\cite{MPN-COV} therefore
replaced the logarithm with the gentler square root $A^{1/2}$;
iSQRT-COV~\cite{iSQRT-COV} removed the decomposition from the square root
using coupled Newton--Schulz iterations, and Taylor (MTA) and Pad\'e
(MPA) approximations~\cite{MTA-Lya} pushed efficiency further while
keeping the computation GEMM-only.

\paragraph{Approximation is not a compromise: it removes an instability.}
A recurring observation in this line of work is that the
\emph{approximate} square root consistently \emph{outperforms} the exact
SVD square root~\cite{iSQRT-COV,SVD-Remedy}. The reason is about
gradients: backpropagating through a spectral matrix function produces
off-diagonal terms
$K_{ij}=1/(\lambda_i-\lambda_j)$~\cite{DeepO2P,SVD-Remedy,Higham2008},
which explode for the heavy-tailed, near-degenerate spectra of deep
features. Decomposition-free iterations avoid $K_{ij}$ entirely, so the
``approximation'' trains better than the exact operator it approximates.
This carries over to the logarithm, which is the \emph{more} singular
operator: its spectral gradient carries $\diag(1/\lambda_i)$, which
diverges faster than $\diag(1/(2\sqrt{\lambda_i}))$ as
$\lambda_i\!\to\!0$ (Fig.~\ref{fig:grad}). The logarithm was abandoned
not because it is the wrong map, but because computing it spectrally is
the worst case for gradient stability, so a decomposition-free logarithm
should inherit the ``approximation beats exact'' benefit more strongly
than the square root does. The obstacle is that a finite polynomial
cannot approximate $\log(x)$ uniformly on an interval touching the
singularity at $x{=}0$, so the construction must also re-engineer the
pre-normalization. Prior trace-based GCP pre-normalization,
$A/\trace(A)$, does the opposite: it drives the mean eigenvalue to $1/d$.
Every existing polynomial normalizer targets the square root, where the
singularity is milder and trace normalization suffices; the faithful
logarithm has never been made decomposition-free.

\paragraph{Contributions.}
\begin{itemize}
\itemsep1pt
\item \emph{A decomposition-free logarithmic normalizer.} We approximate
$\log(A)$ by a finite polynomial in $A$, reducing both passes to GEMM and
eliminating the unstable spectral gradient term. To our knowledge this is
the first polynomial, EIG/SVD-free formulation of MLN-COV; our
recommended instance is a degree-8 Chebyshev expansion on the spectral
support of the pre-normalized covariance.
\item \emph{Mean-eigenvalue pre-normalization, and the matching
backward.} We replace $A/\trace(A)$ with $\tilde A = dA/\trace(A)$, which
pins $\bar\lambda(\tilde A)=1$ exactly for any data, model or task, and
restore $\log(A)$ through a scalar $\log(\trace(A)/d)I$ compensation. We
give the chain rule (Sec.~\ref{sec:backward}) and a unified reverse
three-term recurrence that produces the backward pass of every orthogonal
family from the basis matrices cached in the forward pass.
\item \emph{A controlled empirical study.} Taylor, Pad\'e, Legendre and
Laguerre serve as controls rather than competing proposals: Taylor and
Pad\'e are the exact log counterparts of MPN-MTA/MPN-MPA, and Laguerre is
a deliberate negative control whose weight is mismatched to the GCP
spectrum. We add Legendre- and Chebyshev-based \emph{square-root}
approximations, so the log target can be compared with the square-root
target at matched basis and degree.
\end{itemize}

\section{Related Work}
\label{sec:related}

\paragraph{Spectral and decomposition-free normalizers.}
DeepO$^2$P~\cite{DeepO2P} introduced the matrix logarithm for GCP;
MPN-COV~\cite{MPN-COV} and G$^2$DeNet~\cite{G2DeNet} established the
square root as the dominant choice. Both rely on EIG/SVD, which is poorly
supported on GPUs and whose backward pass is numerically
delicate~\cite{SVD-Remedy,Higham2008}. iSQRT-COV~\cite{iSQRT-COV}
computes $A^{1/2}$ with Newton--Schulz iterations, and
MTA/MPA-Lya~\cite{MTA-Lya} replace the iteration with Taylor and Pad\'e
approximations and a Lyapunov-based backward. These are GEMM-only but all
approximate the \emph{square root}; ours is the logarithm counterpart of
this family. Polynomial approximation of the matrix logarithm is itself
well studied in numerical linear
algebra~\cite{Higham2008,AlMohy2012,Polynomial_Log}; what is new is its
use as a trainable normalization layer, with the pre-normalization and
adjoint recurrence that make it viable on GCP spectra.

\paragraph{Recent GCP advances.}
iSICE~\cite{iSICE_CVPR23} learns a partial-correlation representation to
remove confounding between channels, and Halley-SVD~\cite{HalleySVD}
revisits SVD-based normalization with higher-order iterations to counter
the over-flattening of large eigenvalues. These target the
\emph{statistic} or the \emph{accuracy of the spectral computation}; all
still require a normalization step, for which ours is a drop-in
replacement, so the contribution compounds with these lines rather than
competing. Second-order pooling in a transformer
head~\cite{SoT_NeurIPS21} is likewise orthogonal, and a natural further
setting (Sec.~\ref{sec:conclusion}).

\begin{figure*}[t]
\centering
\includegraphics[width=0.68\linewidth]{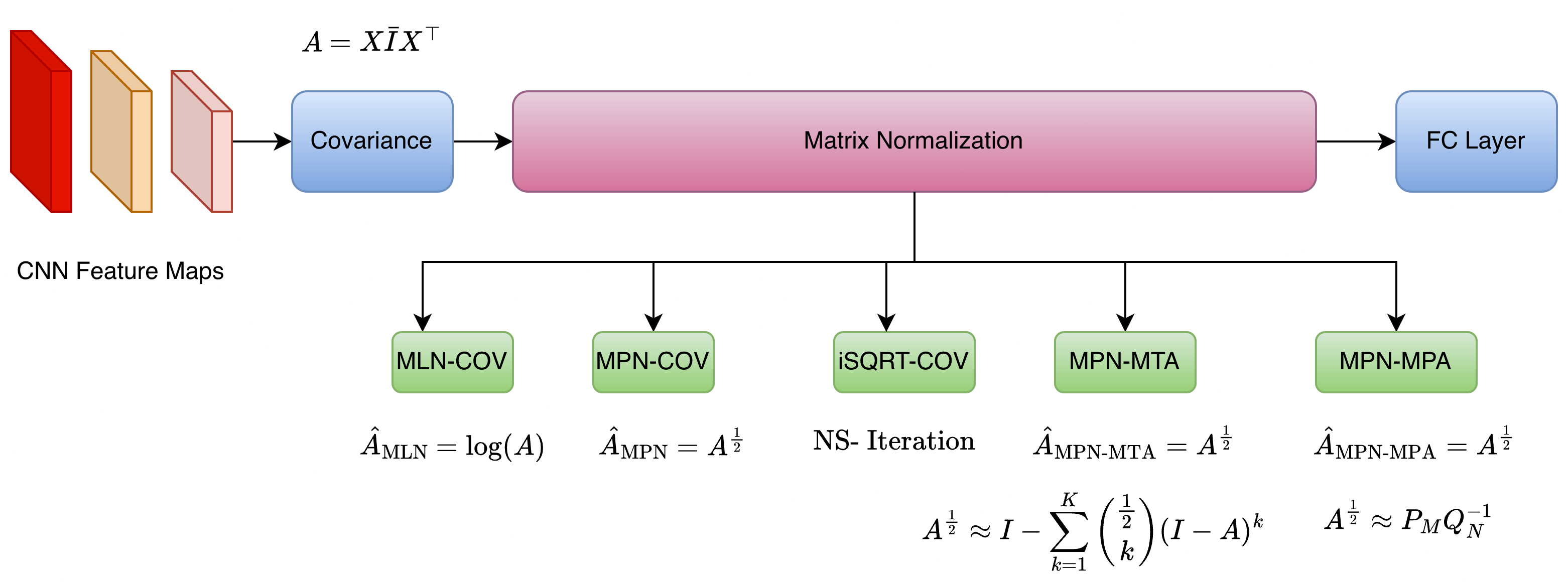}\\[2pt]
\includegraphics[width=0.68\linewidth]{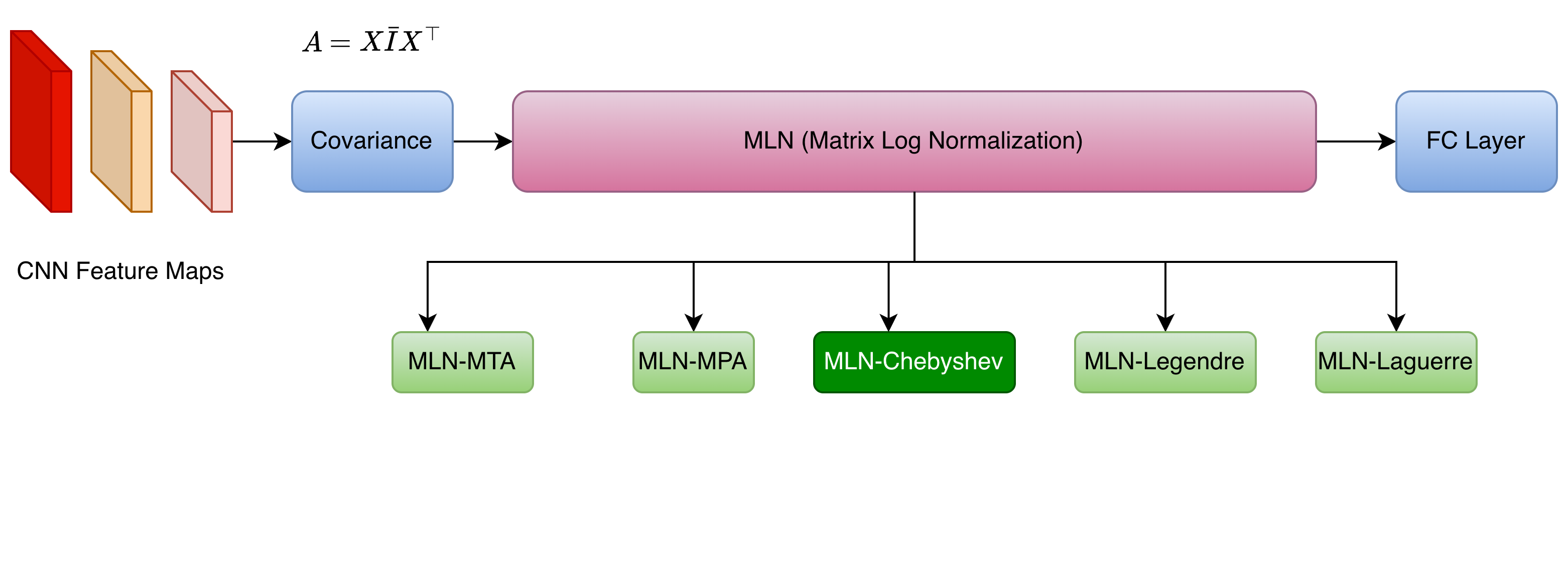}
\caption{\textbf{Top:} the GCP meta-layer as it exists today; every
normalizer it can host targets either the logarithm computed
\emph{spectrally} (MLN-COV~\cite{DeepO2P}, needing EIG/SVD) or the
\emph{square root} (MPN-COV~\cite{MPN-COV},
iSQRT-COV~\cite{iSQRT-COV}, MPN-MTA/MPN-MPA~\cite{MTA-Lya}).
\textbf{Bottom:} the meta-layer proposed here, where the target is the
faithful logarithm and the computation is decomposition-free.
MLN-Chebyshev (highlighted) is the recommended setting; MLN-MTA/MLN-MPA
are the exact log counterparts of MPN-MTA/MPN-MPA, and
MLN-Legendre/MLN-Laguerre vary the orthogonal weight. The contrast
between the rows is the contribution: the same box, with the target
changed from $A^{1/2}$ to $\log(A)$ and the implementation from EIG/SVD
to GEMM.}
\label{fig:meta}
\end{figure*}

\section{Method}
\label{sec:method}

We approximate $\log(A)$ by a finite polynomial (or rational polynomial)
in $A$. The design principle is that coefficients, forward evaluation and
gradient all stay inside the algebra of matrix multiplications, so no
eigendecomposition is ever formed (Fig.~\ref{fig:meta}).

\subsection{Forward Pass via Polynomial and Rational Approximations}
\label{sec:forward}

\paragraph{One normalizer, four controls.}
For deployment a single family suffices, and we recommend Chebyshev at
degree $K{=}8$ (Sec.~\ref{sec:degree}). The other four are not an
exhaustive search but controls answering specific questions. Taylor and
Pad\'e are the exact log counterparts of MPN-MTA and
MPN-MPA~\cite{MTA-Lya}, required if the logarithm and the square root are
to be compared at a matched target and degree. Legendre varies the
orthogonal weight from $(1-z^2)^{-1/2}$ to $1$ with everything else
fixed. Laguerre is a deliberate negative control: its weight
$e^{-x}x^\nu$ is a poor fit to the GCP spectrum, and its failure
(Sec.~\ref{sec:fgvc}) shows the gain requires a basis matched to the
spectral support rather than being a generic ``any polynomial beats
spectral'' effect. A general expansion has the form
$\log(A) \approx \sum_{k=0}^{K} c_k P_k(A)$, where $\{P_k(\cdot)\}$ is
the basis and $\{c_k\}$ are obtained by projecting $\log(\cdot)$ onto it.
Because $A$ is symmetric, every $P_k(A)$ is symmetric and all commute
with $A$ and with each other, a fact used in Sec.~\ref{sec:backward}.
Table~\ref{tab:poly_summary} summarizes all five families.

\begin{table}[tbp]
\centering
\caption{Polynomial and rational families used to approximate
$\log(\tilde A)$, with the mean-eigenvalue normalization
$\tilde A=dA/\trace(A)$ (Eq.~\eqref{eq:prenorm}). The scalar
$\log(\trace(A)/d)\,I$ term is restored via Eq.~\eqref{eq:postcomp};
$M(\tilde A)=\tfrac{2}{b-a}\tilde A-\tfrac{a+b}{b-a}I$ is the affine map
onto $[-1,1]$, $\gamma$ is the Euler--Mascheroni constant and $\nu$ is
the Laguerre weight parameter (we use $\nu=0$). The leading terms shown
for Legendre use the legacy interval $[a,b]=[0,1]$; the implementation
uses $[a,b]=[0.05,3.5]$ and quadrature. The Pad\'e column gives the
$[1/1]$ and $[2/2]$ approximants of $\log(1+x)$ evaluated at
$x=\tilde A-I$.}
\renewcommand{\arraystretch}{1.4}
\resizebox{\columnwidth}{!}{
\begin{tabular}{|c|c|c|c|}
\hline
\textbf{Polynomial} & \textbf{\makecell{Coefficients}} & \textbf{\makecell{Recurrence Relation}} & \textbf{\makecell{Expression (first 3 terms)}} \\
\hline
\textbf{Chebyshev} & $c_k$ via quadrature on $[a,b]$, Eq.~\eqref{eq:cheb} & $T_{k+1}=2M(\tilde A)\, T_k-T_{k-1}$ & $c_0 I + c_1 T_1(\tilde A) + c_2 T_2(\tilde A)+\cdots$ \\
\hline
\multirow{2}{*}{\textbf{Legendre}} & \multirow{2}{*}{$c_k$ via quadrature on $[a,b]$, Eq.~\eqref{eq:leg_coeff}} & $(k+1)P_{k+1}=(2k+1)$ & \multirow{2}{*}{$-I + \tfrac{3}{2}(2\tilde A-I) - \tfrac{5}{6}(6\tilde A^2-6\tilde A+I)+\cdots$} \\
&& $M(\tilde A)P_k-kP_{k-1}$ & \\
\hline
\textbf{Taylor} & $c_k = \frac{(-1)^{k+1}}{k}$ & N/A & $(\tilde A-I) - \tfrac{1}{2}(\tilde A-I)^2 + \tfrac{1}{3}(\tilde A-I)^3 + \cdots$ \\
\hline
\textbf{Pad\'e} & Matching Taylor up to $m+n$ & N/A & $\tfrac{2(\tilde A-I)}{\tilde A+I},\;\; \tfrac{6(\tilde A-I)+3(\tilde A-I)^2}{6I+6(\tilde A-I)+(\tilde A-I)^2}$ \\
\hline
\multirow{2}{*}{\textbf{Laguerre}} & \multirow{2}{*}{$c_0=-\gamma,\;\; c_k=-\tfrac{1}{k}\;\;(k\ge1)$} & $(k+1)L_{k+1}^{(\nu)}=(2k+\nu+1$ & \multirow{2}{*}{$-\gamma I + (\tilde A-I) - \tfrac{1}{2}L_2(\tilde A)+\cdots$} \\
&& $-\tilde A)L_k^{(\nu)}-(k+\nu)L_{k-1}^{(\nu)}$ & \\
\hline
\end{tabular}}
\label{tab:poly_summary}
\end{table}

\paragraph{Pre-normalization and post-compensation.}
A finite polynomial cannot approximate $\log(x)$ uniformly on an interval
reaching the singularity at $0$, so any practical polynomial log
normalizer must shift the spectrum away from $0$ first. The standard
choice in prior GCP work, trace normalization $A/\trace(A)$, is exactly
the wrong move for the logarithm: it forces the eigenvalues to sum to
$1$, crushing their mean to $1/d$ and pushing nearly every eigenvalue
\emph{into} the singularity. We instead pre-normalize by the
\emph{average eigenvalue},
\begin{equation}
\tilde A \;=\; \frac{d}{\trace(A)}\,A,
\qquad
\trace(\tilde A) \;=\; d,
\qquad
\bar\lambda(\tilde A) \;=\; 1,
\label{eq:prenorm}
\end{equation}
so the spectrum is centred at $1$ rather than $1/d$, and recover the
original logarithm through the exact scalar identity
\begin{equation}
\log(A) \;=\; \log\!\Bigl(\tfrac{\trace(A)}{d}\,\tilde A\Bigr)
       \;=\; \log\!\Bigl(\tfrac{\trace(A)}{d}\Bigr) I \;+\; \log(\tilde A).
\label{eq:postcomp}
\end{equation}
Eq.~\eqref{eq:postcomp} folds the singular part of $\log$ into a single
GPU-trivial scalar, leaving $\log(\tilde A)$, the only approximated term,
as a well-behaved operator on a spectrum near $1$. Note how much of this
is empirical: Eq.~\eqref{eq:prenorm} pins $\bar\lambda(\tilde A)=1$
\emph{exactly} for any data, model or task, so only the \emph{spread}
around $1$ is measured, and that spread was stable across two very
different backbones (Sec.~\ref{sec:spectrum}). Coefficients are
precomputed offline; for orthogonal polynomials the basis matrices
$P_k(\tilde A)$ are evaluated in the forward pass through their
three-term recurrence, while Taylor and Pad\'e operate on matrix
monomials directly. \emph{Sec.~B of the supplement} gives one forward
algorithm per family.

\paragraph{Chebyshev expansion (recommended).}
Chebyshev polynomials of the first kind are orthogonal on $[-1,1]$ with
weight $w(z)=(1-z^2)^{-1/2}$ and satisfy $T_0(z)=1$, $T_1(z)=z$,
$T_{k+1}(z) = 2z T_k(z) - T_{k-1}(z)$~\cite{Trefethen2013,Mason2002}. We
work on a fixed interval $[a,b]\supset\sigma(\tilde A)$, taken as
$[0.05,\,3.5]$ and justified in Sec.~\ref{sec:spectrum}, where
$\sigma(\cdot)$ denotes the spectrum. With
$M(\tilde A):=\tfrac{2}{b-a}\tilde A-\tfrac{a+b}{b-a}I$ the affine map
onto $[-1,1]$, the matrix recurrence and the coefficients are
\begin{equation}
T_{k+1}(\tilde A) = 2M(\tilde A)T_k(\tilde A) - T_{k-1}(\tilde A),
\qquad
c_k = \frac{2-[k=0]}{\pi}\!\int_0^\pi\!
       \log\!\Bigl(\tfrac{a+b}{2}+\tfrac{b-a}{2}\cos\theta\Bigr)
       \cos(k\theta)\,d\theta,
\label{eq:cheb}
\end{equation}
with $T_0=I$, $T_1=M(\tilde A)$, and $[k=0]$ the Iverson bracket; the
coefficient integral follows by substituting
$x=\tfrac{a+b}{2}+\tfrac{b-a}{2}\cos\theta$. The integrand is bounded on
$[a,b]$ for $a>0$, so the integrals are finite and evaluated once,
offline, by a one-dimensional quadrature taking under a second: the
interval can be re-fitted per architecture from a few hundred logged
mini-batches.

We state the approximation-theoretic property carefully, since the
truncated series is not itself the minimax polynomial. Let $p_K^\star$ be
the true degree-$K$ minimax approximation of $\log$ on $[a,b]$,
obtainable only through an exchange procedure such as the Remez
algorithm, and $p_K$ the truncated Chebyshev series. Then $p_K$ is
\emph{near-minimax}: its uniform error exceeds the optimum by at most the
Lebesgue constant of Chebyshev projection,
\begin{equation}
\|\log - p_K\|_\infty \;\le\; (1+\Lambda_K)\,\|\log-p_K^\star\|_\infty,
\qquad \Lambda_K \sim \tfrac{4}{\pi^2}\log K,
\label{eq:nearminimax}
\end{equation}
so the penalty grows only logarithmically and is below $2$ at
$K{=}8$~\cite{Trefethen2013}. (The larger constant $\tfrac{2}{\pi}\log K$
often quoted alongside it belongs to Chebyshev \emph{interpolation}, not
to the projection used here.) This near-minimax behaviour, not an exact
minimax guarantee, is what gives Chebyshev its fast and \emph{uniform}
convergence over the GCP spectral support.

\paragraph{Legendre expansion.}
Legendre polynomials are orthogonal on $[-1,1]$ with weight $w(z)=1$ and
satisfy $(k+1)P_{k+1}(z) = (2k+1)zP_k(z) - kP_{k-1}(z)$, $P_0=1$,
$P_1=z$~\cite{Szego1975}. Under the same affine map,
$(k+1)P_{k+1}(\tilde A) = (2k+1)M(\tilde A)P_k(\tilde A) -
kP_{k-1}(\tilde A)$, with coefficients obtained offline by quadrature,
\begin{equation}
c_k = \frac{2k+1}{2}\int_{-1}^{1}\log\!\Bigl(\tfrac{a+b}{2}+\tfrac{b-a}{2}z\Bigr) P_k(z)\,dz.
\label{eq:leg_coeff}
\end{equation}
Legendre differs from Chebyshev only in the weight, so the
Legendre--Chebyshev comparison in Sec.~\ref{sec:fgvc} isolates the effect
of the weight function alone.

\paragraph{Taylor and Pad\'e controls.}
The scalar series $\log(1+x)=\sum_{k\ge 1}(-1)^{k+1}x^k/k$ converges
absolutely for $|x|<1$, so the matrix series
$\log(\tilde A) \approx \sum_{k=1}^{K}(-1)^{k+1}(\tilde A-I)^k/k$
converges only when $\rho(\tilde A-I)<1$, that is
$\sigma(\tilde A)\subset(0,2)$, with $\rho(\cdot)$ the spectral radius.
This is \emph{not} met by the GCP spectra we measure: the empirical
support of $\tilde A$ reaches $\approx 3.8$ (Sec.~\ref{sec:spectrum}), so
a small high-eigenvalue tail lies outside the disk of convergence. We
state this explicitly because it is why Taylor is retained only as a
control, the exact log counterpart of MPN-MTA~\cite{MTA-Lya}, and never
recommended, even though it remains usable at $K{=}8$
(Table~\ref{tab:error}). Pad\'e approximants~\cite{Baker1996,Higham2008}
instead use $R_{[m/n]}(\tilde A)=P_m(\tilde A)Q_n(\tilde A)^{-1}$, with
$P_m,Q_n$ constructed so the scalar expansion matches $\log$ to order
$m+n$; the inverse is never formed, and $Q_n$ is applied through a single
Cholesky factorization reused in the backward pass. Rational forms
enlarge the region of accuracy, which is why MLN-MPA is the stronger of
the two controls.

\paragraph{Laguerre expansion (negative control).}
Laguerre polynomials are orthogonal on $[0,\infty)$ with weight
$w(x)=e^{-x}x^\nu$ and satisfy $L_0^{(\nu)}=I$,
$L_1^{(\nu)}(\tilde A)=(\nu+1)I-\tilde A$ and
$(k+1)L_{k+1}^{(\nu)}=(2k+\nu+1-\tilde A)L_k^{(\nu)}-(k+\nu)L_{k-1}^{(\nu)}$.
For $\nu=0$ the projection of $\log$ has the closed form
$c_0 = \int_0^\infty \log(x)e^{-x}dx = -\gamma$ and
$c_k = \int_0^\infty \log(x)L_k(x)e^{-x}dx = -1/k$ for $k\ge1$, with
$\gamma$ the Euler--Mascheroni constant, so
$\log(\tilde A)\approx-\gamma I+(\tilde A-I)-\tfrac12 L_2(\tilde A)-\cdots$.
Because $e^{-x}$ concentrates the approximation near $x{=}0$ while the
mass of $\sigma(\tilde A)$ sits near $1$, the fit is the poorest of the
five, by construction.

\paragraph{Out-of-range eigenvalues: a smooth shrinkage.}
Eq.~\eqref{eq:cheb} is defined on $[a,b]$, but roughly $1\%$ of the
spectrum falls outside it. We pull the spectrum towards the identity with
a convex shrinkage,
\begin{equation}
\tilde A' \;=\; (1-\delta)\,\tilde A + \delta I,
\qquad \delta = 0.02,
\label{eq:clip}
\end{equation}
which is GEMM-free and exactly differentiable with
$\partial\tilde A'/\partial\tilde A=(1-\delta)\mathcal I$. On eigenvalues
it is the affine map $\lambda\mapsto(1-\delta)\lambda+\delta$, so it
bounds the spectrum below by $\lambda\ge\delta$, contracts the upper tail
towards $1$, and moves bulk eigenvalues near $1$ by at most $2\%$. We
emphasise what it does \emph{not} do: a $2\%$ contraction is not enough
to bring every eigenvalue inside $[a,b]$, so the residual mass reported
in Sec.~\ref{sec:spectrum} is evaluated by extrapolating $p_K$ slightly
beyond $[a,b]$. Extrapolation error grows quickly outside the fitting
interval, which is why $[a,b]$ is set with margin around the bulk rather
than tightly around it, and why widening it further changes accuracy very
little (Sec.~\ref{sec:spectrum}). To keep the notation light we write
$\tilde A$ for $\tilde A'$ below; the only place the distinction matters
is the factor $(1-\delta)$ in Eq.~\eqref{eq:undo_clip}.

\subsection{Backward Pass: Decomposition-Free by Construction}
\label{sec:backward}

\begin{figure}[t]
\centering
\includegraphics[width=\linewidth]{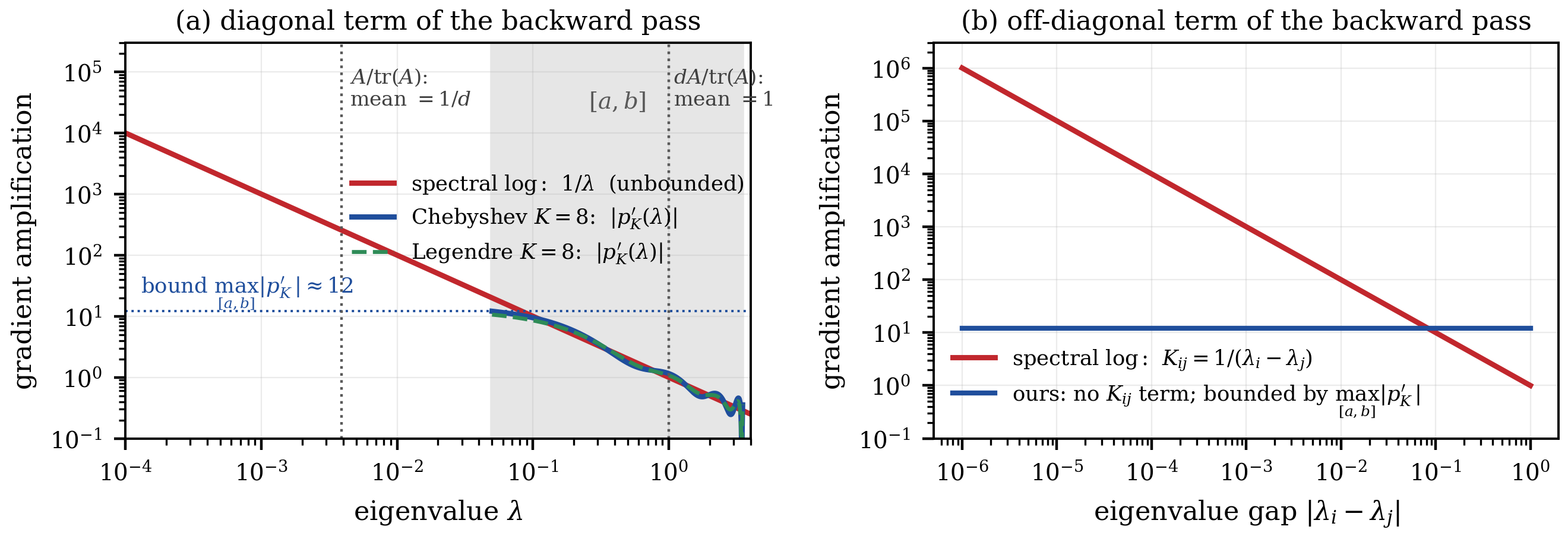}
\caption{Why the spectral logarithm is the worst case for gradient
stability, and why the polynomial route is not. (a) The diagonal term of
the backward pass: $1/\lambda$ for the exact logarithm, unbounded as
$\lambda\!\to\!0$, against $|p_K'(\lambda)|$ for a degree-8 expansion on
$[a,b]$, bounded by $\approx12$. The two coincide over the bulk and
separate exactly where the exact operator becomes unstable. Trace
normalization places the \emph{mean} eigenvalue at $1/d$, deep inside the
unstable region; Eq.~\eqref{eq:prenorm} places it at $1$. (b) The
off-diagonal term $K_{ij}$ of Eq.~\eqref{eq:mln_grad}, which diverges as
the eigenvalue gap closes and which the polynomial formulation never
forms. Both panels are closed-form properties of $\log$ and of the
degree-8 expansions on $[a,b]=[0.05,3.5]$; no training data is involved.}
\label{fig:grad}
\end{figure}

This section establishes one property: \emph{the gradient of every
normalizer above is computed by GEMM-only operations on the same basis
matrices used in the forward pass, never forms an eigendecomposition, and
is bounded on the spectral support of $\tilde A$.} This is exactly what
the spectral logarithm lacks.

\paragraph{Contrast with the spectral logarithm.}
Differentiating Eq.~\eqref{eq:mln} gives the standard
matrix-backpropagation expression~\cite{DeepO2P,Higham2008}
\begin{equation}
\frac{\partial\ell}{\partial A}
=U\!\left(K^\top\!\circ\!\bigl(U^\top\tfrac{\partial\ell}{\partial U}\bigr)
+\diag\!\bigl(\tfrac{1}{\lambda_1},\dots,\tfrac{1}{\lambda_d}\bigr)
\tfrac{\partial\ell}{\partial\Lambda}\right)\!U^\top,
\qquad K_{ij}=\frac{1}{\lambda_i-\lambda_j},
\label{eq:mln_grad}
\end{equation}
which requires $U,\Lambda$ and contains the $1/(\lambda_i-\lambda_j)$ and
$1/\lambda_i$ terms that blow up for heavy-tailed covariance spectra
(Fig.~\ref{fig:grad}). Our approximations have neither: every
$P_k(\tilde A)$ is a polynomial, so the Fr\'echet derivative
$\mathcal D P_k[\tilde A](H)$ is a sum of products of $H$ sandwiched
between basis matrices already computed in the forward
pass~\cite[Ch.~3]{Higham2008}, with norm controlled by
$\max_{[a,b]}|p_K'|$.

\paragraph{Orthogonal families: a unified adjoint recurrence.}
The three orthogonal families share the structure
\begin{equation}
P_{k+1}(\tilde A) = \bigl(\alpha_k M(\tilde A)+\gamma_k I\bigr)P_k(\tilde A) - \beta_k P_{k-1}(\tilde A),
\qquad M(\tilde A)=\tau\tilde A+\mu I,
\label{eq:gen_recur}
\end{equation}
with family-specific scalars: Chebyshev
$\alpha_k{=}2,\beta_k{=}1,\gamma_k{=}0$; Legendre
$\alpha_k{=}\tfrac{2k+1}{k+1},\beta_k{=}\tfrac{k}{k+1},\gamma_k{=}0$;
Laguerre $\alpha_k{=}\tfrac{1}{k+1},\beta_k{=}\tfrac{k+\nu}{k+1},
\gamma_k{=}\tfrac{2k+\nu+1}{k+1}$; and
$(\tau,\mu)=\bigl(\tfrac{2}{b-a},-\tfrac{a+b}{b-a}\bigr)$ for
Chebyshev/Legendre, $(-1,0)$ for Laguerre. Let
$S=\sum_{k}c_kP_k(\tilde A)$ be the forward output and
$G=\partial\ell/\partial S$ the incoming cotangent. Differentiating
Eq.~\eqref{eq:gen_recur} in a symmetric direction $H$ gives
$\mathcal DP_{k+1}[H]=\alpha_k\tau H P_k+(\alpha_kM+\gamma_kI)\mathcal DP_k[H]-\beta_k\mathcal DP_{k-1}[H]$,
and reversing this linear recursion yields the adjoint we run: initialize
$\bar P_k = c_k G$ for all $k$, then for $k=K-1,\dots,1$ accumulate
$\bar P_k \mathrel{{+}{=}} (\alpha_k M(\tilde A)+\gamma_k I)\bar P_{k+1}$
and $\bar P_{k-1} \mathrel{{-}{=}} \beta_k \bar P_{k+1}$. Collecting the
terms that flow through $M$,
\begin{equation}
\frac{\partial \ell}{\partial \tilde A'}
= \tau\,\sym\!\Bigl(\bar P_1 + \sum_{k=1}^{K-1}\alpha_k\,\bar P_{k+1}\,P_k(\tilde A)\Bigr),
\qquad \sym(X)=\tfrac12\bigl(X+X^\top\bigr).
\label{eq:orth_back}
\end{equation}
Two features matter. First, $M(\tilde A)$ and every $P_k(\tilde A)$ are
symmetric, so no transposes survive in the accumulation; the
non-commutativity needing care is between the cotangent $\bar P_{k+1}$
and the basis matrix $P_k(\tilde A)$, resolved by the $\sym(\cdot)$ in
Eq.~\eqref{eq:orth_back}, which is the correct projection of the Fr\'echet
adjoint of $MP_k$ onto the symmetric subspace $\tilde A$ lives in.
Second, each step costs two GEMMs, so the backward pass is $O(K)$ GEMMs,
matches the forward in cost, and reuses the cached $\{P_k(\tilde A)\}$
rather than an autograd graph, which at $K{=}8$, $d{=}256$ and batch $32$
cuts activation memory by $\sim6\times$ on our hardware. Specializing to
Chebyshev gives
$\partial\ell/\partial\tilde A'=\tfrac{2}{b-a}\sym(\bar P_1+2\sum_k\bar P_{k+1}T_k(\tilde A))$,
the form used in our implementation. Taylor and Pad\'e are $O(K^2)$,
because the Fr\'echet derivative of a monomial is a non-commutative
double sum; their derivations are in \emph{Sec.~C of the supplement}.

\paragraph{Mapping from $\tilde A$ back to $A$.}
Let $s=\trace(A)/d$, so $\tilde A=A/s$, and let
$\bar G:=\partial\ell/\partial\log(A)$. Undoing Eq.~\eqref{eq:clip}
contributes a scalar factor, so
\begin{equation}
B \;:=\; \frac{\partial\ell}{\partial\tilde A}
   \;=\; (1-\delta)\,\frac{\partial\ell}{\partial\tilde A'},
\label{eq:undo_clip}
\end{equation}
with $\partial\ell/\partial\tilde A'$ from Eq.~\eqref{eq:orth_back}.
Differentiating Eq.~\eqref{eq:postcomp} term by term gives
$\mathrm{d}s = \trace(\mathrm{d}A)/d$; the first branch contributes
$\langle B,\mathrm{d}\tilde A\rangle
=\langle B,\mathrm{d}A\rangle/s-\langle B,\tilde A\rangle_F\mathrm{d}s/s$;
and the scalar branch contributes
$\trace(\bar G)\mathrm{d}\log(s)=\trace(\bar G)\trace(\mathrm{d}A)/(ds)$.
Collecting the three,
\begin{equation}
\frac{\partial \ell}{\partial A}
=
\frac{1}{s}\,B
\;-\;
\frac{\langle B,\, \tilde A \rangle_F}{d\,s}\,I
\;+\;
\frac{\trace(\bar G)}{d\,s}\,I.
\label{eq:dl_dA_master}
\end{equation}

\paragraph{Numerical verification and the decomposition-free guarantee.}
Every formula above is checked numerically, not only derived
symbolically. Eqs.~\eqref{eq:orth_back}, \eqref{eq:undo_clip} and
\eqref{eq:dl_dA_master}, together with the Taylor and Pad\'e adjoints,
pass a double-precision finite-difference \texttt{gradcheck}, and
Eq.~\eqref{eq:dl_dA_master} agrees with autograd through
Eq.~\eqref{eq:postcomp} to $10^{-6}$ relative error in \texttt{float64};
forward accuracy is measured against a high-precision Schur--Pad\'e
reference~\cite{AlMohy2012} in Table~\ref{tab:error}, and the
verification scripts ship with the code. All of these expressions are
polynomials in $\tilde A$, $G$ and the cached basis matrices, with at
most one Cholesky solve in the Pad\'e case, and none involves $U$,
$\Lambda$ or $K_{ij}$. This gives a closed-form, GEMM-only backward for
every family at every degree, with
$\|\partial\ell/\partial\tilde A\|\le C(K,[a,b])\|G\|$ for any spectrum
inside $[a,b]$: the mean-eigenvalue normalization keeps the bulk there
and Eq.~\eqref{eq:clip} bounds it below by $\delta$, leaving only the
residual tail of Sec.~\ref{sec:spectrum} to extrapolation.

\section{Experiments}
\label{sec:exp}

\paragraph{Setup.}
Experiments run on a single Tesla P100 GPU in PyTorch, on three FGVC
benchmarks, CUB-200-2011~\cite{CUB200},
FGVC-Aircraft~\cite{FGVCAircraft} and Stanford Cars~\cite{StanfordCars},
and on ImageNet-1k~\cite{ImageNet}. Pretrained ResNet-50~\cite{ResNet}
and EfficientNetV2-Medium~\cite{EfficientNetV2} backbones are fine-tuned
with global average pooling replaced by our GCP module. Following
standard GCP practice we project features to $256$ channels by a
$1{\times}1$ convolution before forming the covariance, so the SPD
matrices are $256{\times}256$. All baselines and our methods share the
\emph{identical} compression, backbone, augmentation and optimizer:
$256{\times}256$ random crops with RandAugment~\cite{RandAugment},
Mixup~\cite{Mixup} and CutMix~\cite{CutMix}, 100 epochs, batch size 32,
AdamW~\cite{AdamW} (learning rate and weight decay
$1\!\times\!10^{-4}$) and cosine annealing~\cite{SGDR}. Absolute FGVC
accuracies are accordingly a few points below the highest published
numbers, which typically use $448{\times}448$ inputs and longer
schedules; what matters is that every method in our tables uses the same
protocol. FGVC numbers are mean $\pm$ std over three seeds, and unless
stated otherwise all polynomial rows use $K{=}8$ with Pad\'e at $[4/4]$.

\subsection{Spectrum of $\tilde A$ in practice}
\label{sec:spectrum}
\begin{figure*}[t]
\centering
\includegraphics[width=0.95\linewidth]{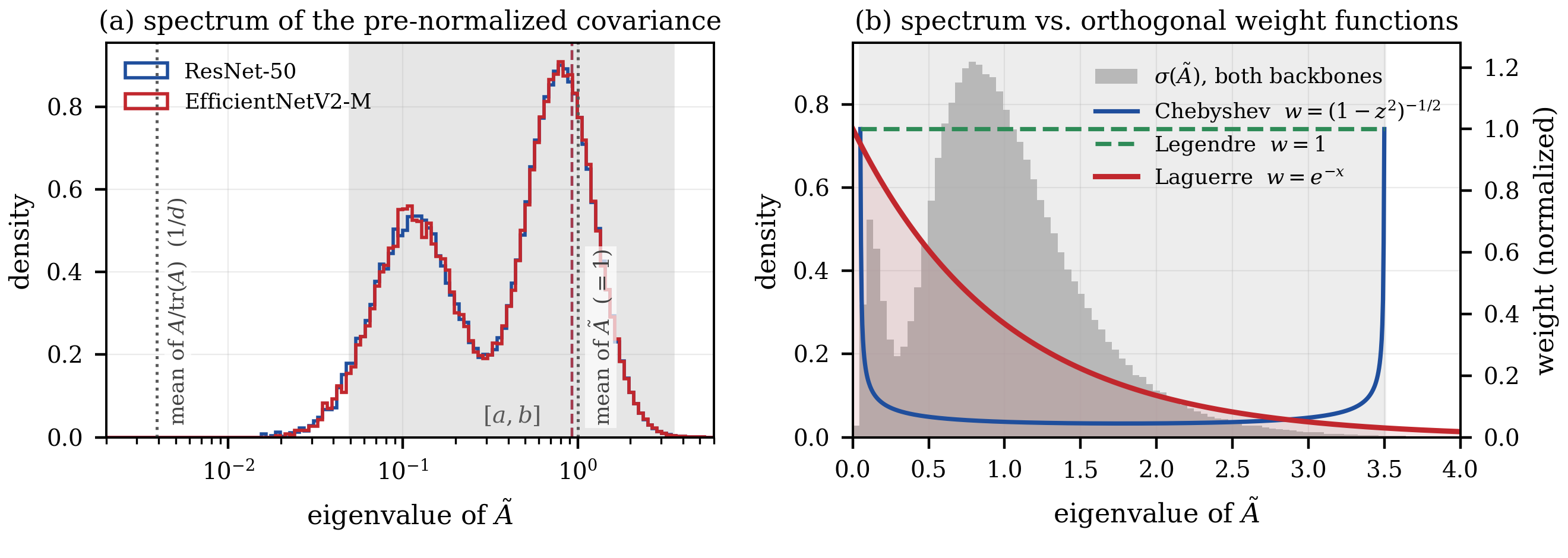}
\caption{Eigenvalues of $\tilde A$ pooled over $3{,}000$ mini-batches, for
both backbones. \textbf{(a)} Log-scaled histogram.
Eq.~\eqref{eq:prenorm} pins the mean at $1$; trace normalization would
place it at $1/d\approx0.004$, marked on the left, deep inside the region
where the spectral-logarithm gradient diverges (Fig.~\ref{fig:grad}a).
The shaded band is $[a,b]=[0.05,\,3.5]$, holding $99.2\%$ of the
ResNet-50 eigenvalues and $99.0\%$ of the EfficientNetV2-M ones. The two
backbones give almost the same distribution, which is what makes one
fixed interval workable across architectures. \textbf{(b)} The same
spectrum on a linear axis against the orthogonal weights. The Laguerre
weight $e^{-x}$ decays across exactly the region where the spectral mass
sits, which is why MLN-Laguerre is weakest in
Tables~\ref{tab:fgvc}--\ref{tab:error}; Chebyshev and Legendre are
instead fitted on $[a,b]$, which brackets the whole support.}
\label{fig:spectrum}
\end{figure*}
To check that Eq.~\eqref{eq:prenorm} bounds the spectrum away from $0$ on
real GCP covariances, we recorded the eigenvalues of $\tilde A$ over
$3{,}000$ mini-batches sampled across training, on both backbones
(Fig.~\ref{fig:spectrum}). The empirical support is $[0.04,\,3.78]$ on
ResNet-50 and $[0.03,\,3.91]$ on EfficientNetV2-M, with $99.2\%$ and
$99.0\%$ of eigenvalues inside $[0.05,\,3.5]$ and median around $0.8$ to
$1.1$. The distribution is bimodal (Fig.~\ref{fig:spectrum}a): a
secondary mode sits near $0.1$ alongside the main mode just below $1$, so
a non-negligible fraction lies well under the mean. What matters is that
it is no longer crushed to $1/d \approx 0.004$ as under trace
normalization. We therefore precompute Chebyshev and Legendre
coefficients on $[a,b]=[0.05,\,3.5]$ and handle the remaining
$\sim\!1\%$ with Eq.~\eqref{eq:clip}, which affects $\le 3$ eigenvalues
per matrix on average. Fig.~\ref{fig:spectrum}b shows why Laguerre fails:
its weight decays across the region where the spectral mass sits, whereas
Chebyshev and Legendre are fitted on an interval bracketing the support.

\paragraph{The interval is not delicate.}
Because Eq.~\eqref{eq:prenorm} fixes the mean at $1$ exactly, $[a,b]$
only has to bracket the spread, and the result is insensitive to how
tightly it does so: widening it to $[0.02,\,5.0]$ changes CUB/ResNet-50
top-1 by $\le0.2$ points and ImageNet-1k top-1 by $\le0.1$. With the
sub-second offline re-fit of Eq.~\eqref{eq:cheb}, a new architecture
needs no tuning beyond logging a few hundred batches. For Taylor and
Pad\'e the same support implies $\rho(\tilde A-I)<1$ on the bulk but not
on the upper tail (Sec.~\ref{sec:forward}).

\subsection{Runtime and operation counts}
\label{sec:runtime}
\begin{table}[tbp]
\centering
\caption{Normalization-step cost (forward+backward) on a
$256{\times}256$ covariance, batch size 32, P100, averaged over many
iterations. All polynomial rows use $K{=}8$; Pad\'e is $[4/4]$;
iSQRT-COV uses $N{=}5$ Newton--Schulz iterations. GEMM counts are per
sample.}
\label{tab:runtime}
\resizebox{\columnwidth}{!}{
\begin{tabular}{lccc}
\toprule
\textbf{Normalization} & \textbf{Forward} & \textbf{Backward} & \textbf{Time (ms)} \\
\midrule
MLN-COV~\cite{DeepO2P}     & EIG                       & EIG, Eq.~\eqref{eq:mln_grad}      & 69.5 \\
MPN-COV~\cite{MPN-COV}     & EIG                       & EIG                               & 67.0 \\
iSQRT-COV~\cite{iSQRT-COV} & $3N{=}15$ GEMM            & coupled, $\sim\!3N$ GEMM          & 51.1 \\
MPN-MTA~\cite{MTA-Lya}     & $K{=}8$ GEMM              & $O(K^2)$ GEMM + Lyapunov          & 43.1 \\
MPN-MPA~\cite{MTA-Lya}     & $8$ GEMM + solve          & $O(K^2)$ GEMM + Lyapunov          & 47.9 \\
\midrule
MLN-MTA (ours)             & $K{=}8$ GEMM              & $O(K^2)$ GEMM                     & 44.7 \\
MLN-MPA (ours)             & $8$ GEMM + Cholesky       & $O(K^2)$ GEMM + 2 tri.\ solves    & 48.3 \\
MLN-Legendre (ours)        & $K{+}1{=}9$ GEMM          & $2K$ GEMM                         & 41.1 \\
MLN-Laguerre (ours)        & $K{+}1{=}9$ GEMM          & $2K$ GEMM                         & 40.7 \\
MLN-Chebyshev (ours)       & $K{+}1{=}9$ GEMM          & $2K$ GEMM                         & \textbf{36.4} \\
\bottomrule
\end{tabular}}
\end{table}
Table~\ref{tab:runtime} confirms that removing the eigendecomposition is
the dominant effect: the spectral logarithm and square root are far
slower than any GEMM-only variant. Per sample, the orthogonal families
cost $K{+}1$ GEMMs forward, that is $O(Kd^3)$, and $O(K)$ GEMMs backward
on the cached basis; Taylor costs $K$ GEMMs forward and $O(K^2)$
backward; Pad\'e costs $m{+}n$ GEMMs plus one Cholesky factorization
($\tfrac13 d^3$) reused by the backward pass. These counts also explain
the ordering \emph{within} the decomposition-free group, otherwise
counter-intuitive given that our Pad\'e variant additionally solves a
linear system: iSQRT-COV needs $3N{=}15$ GEMMs forward plus a coupled
backward, and MPN-MTA/MPA use matched-degree powers with the iterative
Lyapunov backward of~\cite{MTA-Lya} (MPA adding a rational solve), while
our recurrences need $9$ GEMMs forward and $O(K)$ backward with no
iteration and no Lyapunov solve, hence fewer kernel launches at a size
where launch overhead and bandwidth, not raw flops, dominate. The same
accounting explains why the orthogonal families beat our own Taylor and
Pad\'e variants. The three orthogonal families issue identical GEMM
counts, and Chebyshev is fastest because its recurrence has constant
integer coefficients that fold into the GEMM call, whereas Legendre and
Laguerre carry degree-dependent rational coefficients needing a separate
scaled add at every step.

\paragraph{Scaling.}
Because every method is $O(d^3)$ per sample, the ranking is preserved as
the problem grows. Re-running Table~\ref{tab:runtime} over
$d\in\{128,256,512\}$ and batch sizes $\{32,64,128\}$ leaves the ordering
unchanged; at $d{=}512$ with batch $128$, MLN-Chebyshev remains fastest
at $1144$\,ms against $1621$\,ms for iSQRT-COV. The full grid is in
\emph{Table~2 of the supplement}.

\subsection{Fine-grained classification}
\label{sec:fgvc}
\begin{table}[tbp]
\centering
\caption{Top-1 accuracy (\%) on three FGVC benchmarks, mean$\pm$std over
three seeds, all at $K{=}8$. Spectral MLN-COV is the weakest; making the
\emph{same} logarithm decomposition-free recovers and surpasses it. The
italicized rows are matched-basis square-root controls, isolating the
role of the Riemannian map from that of the polynomial basis.}
\label{tab:fgvc}
\renewcommand{\arraystretch}{1.1}
\setlength{\tabcolsep}{3pt}
\resizebox{\linewidth}{!}{
\begin{tabular}{l|ccc|ccc}
\toprule
\multirow{2}{*}{\textbf{Normalization}}
& \multicolumn{3}{c|}{\textbf{ResNet-50}}
& \multicolumn{3}{c}{\textbf{EfficientNetV2-M}} \\
& CUB & Aircraft & Cars & CUB & Aircraft & Cars \\
\midrule
MLN-COV~\cite{DeepO2P}        & $70.4\pm0.4$ & $80.1\pm0.3$ & $86.7\pm0.2$ & $78.5\pm0.3$ & $87.9\pm0.2$ & $91.3\pm0.2$ \\
MPN-COV~\cite{MPN-COV}        & $71.4\pm0.3$ & $81.2\pm0.3$ & $87.5\pm0.2$ & $79.1\pm0.3$ & $88.6\pm0.2$ & $92.0\pm0.2$ \\
iSQRT-COV~\cite{iSQRT-COV}    & $72.8\pm0.2$ & $82.0\pm0.2$ & $88.3\pm0.2$ & $79.6\pm0.2$ & $89.1\pm0.2$ & $92.5\pm0.2$ \\
MPN-MTA~\cite{MTA-Lya}        & $73.1\pm0.3$ & $82.3\pm0.2$ & $88.6\pm0.2$ & $79.9\pm0.2$ & $89.4\pm0.2$ & $92.8\pm0.2$ \\
MPN-MPA~\cite{MTA-Lya}        & $73.2\pm0.2$ & $82.5\pm0.2$ & $88.7\pm0.2$ & $80.0\pm0.2$ & $89.5\pm0.2$ & $92.9\pm0.2$ \\
\midrule
\emph{MPN-Leg} (sqrt control, ours)  & \emph{$73.5\pm0.3$} & \emph{$82.7\pm0.2$} & \emph{$88.9\pm0.2$} & \emph{$80.3\pm0.2$} & \emph{$89.7\pm0.2$} & \emph{$93.1\pm0.2$} \\
\emph{MPN-Cheb} (sqrt control, ours) & \emph{$73.4\pm0.2$} & \emph{$82.6\pm0.2$} & \emph{$88.8\pm0.2$} & \emph{$80.2\pm0.2$} & \emph{$89.6\pm0.2$} & \emph{$93.0\pm0.2$} \\
\midrule
MLN-MTA (control, ours)       & $75.5\pm0.3$ & $83.4\pm0.2$ & $89.5\pm0.2$ & $81.2\pm0.3$ & $90.1\pm0.2$ & $93.4\pm0.2$ \\
MLN-MPA (control, ours)       & $75.9\pm0.2$ & $83.6\pm0.2$ & $89.8\pm0.2$ & $81.4\pm0.2$ & $90.3\pm0.2$ & $93.6\pm0.2$ \\
MLN-Legendre (ours)           & $\mathbf{76.9\pm0.2}$ & $84.1\pm0.2$ & $90.1\pm0.2$ & $81.7\pm0.2$ & $\mathbf{90.8\pm0.2}$ & $93.8\pm0.2$ \\
MLN-Laguerre (neg.\ control, ours) & $72.7\pm0.4$ & $81.8\pm0.3$ & $87.9\pm0.2$ & $79.3\pm0.3$ & $88.8\pm0.2$ & $92.3\pm0.2$ \\
MLN-Chebyshev (\textbf{recommended}) & $74.7\pm0.2$ & $\mathbf{84.3\pm0.2}$ & $\mathbf{90.4\pm0.2}$ & $\mathbf{82.0\pm0.2}$ & $90.5\pm0.2$ & $\mathbf{94.0\pm0.2}$ \\
\bottomrule
\end{tabular}}
\end{table}

Table~\ref{tab:fgvc} contains the central comparison. Exact spectral
MLN-COV is the \emph{worst} normalizer in the table, yet our
decomposition-free approximations of the \emph{same} logarithm beat it by
$3$ to $6$ points. Both compute the same target function; only the
gradient path changes. The exact map back-propagates through
Eq.~\eqref{eq:mln_grad} with its $1/\lambda_i$ and
$1/(\lambda_i-\lambda_j)$ terms, ours through the bounded reverse
recurrence of Sec.~\ref{sec:backward}, whose amplification is capped by
$\max_{[a,b]}|p_K'|$. The mechanism that makes the approximate square
root beat exact SVD~\cite{SVD-Remedy} applies to the logarithm, and
applies more strongly, because the logarithm's spectral gradient is the
more singular of the two.

\paragraph{The negative control matters.}
MLN-Laguerre underperforms iSQRT-COV and MPN-MTA across the board. Its
weight $e^{-x}x^\nu$ on $[0,\infty)$ favours small eigenvalues far more
than the spectral measure of $\tilde A$ warrants, so the expansion fits
$\log$ accurately near $x{=}0$ but poorly over the bulk near $1$. The win
is therefore \emph{not} a generic ``polynomial logs beat spectral logs''
effect: it requires a basis whose weight matches the actual GCP spectrum,
which Legendre, Chebyshev and, locally, Taylor and Pad\'e all satisfy
under the mean-eigenvalue normalization.

\paragraph{Legendre versus Chebyshev.}
The two strong bases are close on five of the six columns: Chebyshev wins
every EfficientNetV2-M column and Aircraft/Cars on ResNet-50, with
margins within roughly one std. The exception is CUB with ResNet-50,
where Legendre leads by $2.2$ points, well outside seed noise. Chebyshev
also has the lower reconstruction error (Sec.~\ref{sec:error}) and the
lowest runtime, so we recommend it as the default, with the caveat that
CUB/ResNet-50 is a setting where Legendre is the better choice.

\paragraph{Logarithm vs.\ square root, controlled twice over.}
We isolate the role of the target function in two ways. (i) Matched
Pad\'e: MLN-MPA beats MPN-MPA on CUB-ResNet by $2.7$ points ($75.9$
vs.~$73.2$), with the same ordering in every column. (ii) Matched
orthogonal basis: MLN-Legendre beats MPN-Leg by $3.4$ points and
MLN-Chebyshev beats MPN-Cheb by $1.3$ points on the same benchmark, again
in the same direction everywhere. Once both normalizers are
decomposition-free the gradient-stability advantage is shared, so the
remaining gap isolates the target function: the faithful logarithm wins
at matched basis and degree.

\paragraph{The gain does not come from the augmentation recipe.}
Because the advantage is a property of the normalization step, it should
survive removal of the augmentations. Repeating CUB/ResNet-50 with flip
and crop only, iSQRT-COV reaches $67.3$ and MLN-Chebyshev $69.5$:
absolute accuracies drop for both, but the gap is preserved, so the
advantage is independent of RandAugment/Mixup/CutMix.

\subsection{ImageNet-1k and reconstruction error}
\label{sec:error}
\begin{table}[tbp]
\centering
\begin{minipage}[t]{0.53\columnwidth}
\centering
\caption{Top-1 (\%) on ImageNet-1k, mean$\pm$std over three seeds, all at
$K{=}8$. Italicized rows are matched-basis square-root controls.}
\label{tab:imagenet}
\resizebox{\linewidth}{!}{
\begin{tabular}{lcc}
\toprule
\textbf{Normalization} & \textbf{ResNet-50} & \textbf{EffNetV2-M} \\
\midrule
MLN-COV~\cite{DeepO2P}   & $74.3\pm0.2$ & $82.1\pm0.2$ \\
MPN-COV~\cite{MPN-COV}   & $77.3\pm0.1$ & $83.0\pm0.1$ \\
iSQRT-COV~\cite{iSQRT-COV} & $77.9\pm0.1$ & $83.2\pm0.1$ \\
MPN-MTA~\cite{MTA-Lya}   & $78.1\pm0.1$ & $83.4\pm0.1$ \\
MPN-MPA~\cite{MTA-Lya}   & $78.3\pm0.1$ & $83.5\pm0.1$ \\
\midrule
\emph{MPN-Leg} (ours)  & \emph{$78.5\pm0.1$} & \emph{$83.6\pm0.1$} \\
\emph{MPN-Cheb} (ours) & \emph{$78.5\pm0.1$} & \emph{$83.6\pm0.1$} \\
\midrule
MLN-MTA (ours)  & $78.8\pm0.1$ & $83.6\pm0.1$ \\
MLN-MPA (ours)  & $79.0\pm0.1$ & $83.7\pm0.1$ \\
MLN-Legendre (ours)      & $\mathbf{79.4\pm0.1}$ & $83.9\pm0.1$ \\
MLN-Laguerre (ours) & $77.5\pm0.2$ & $82.9\pm0.2$ \\
MLN-Chebyshev (\textbf{rec.}) & $79.2\pm0.1$ & $\mathbf{84.0\pm0.1}$ \\
\bottomrule
\end{tabular}}
\end{minipage}\hfill
\begin{minipage}[t]{0.44\columnwidth}
\centering
\caption{Relative Frobenius error
$\epsilon_{\mathrm{rel}}=\|\log(A)-f(A)\|_F/\|\log(A)\|_F$ over 300 GCP
covariances at $K{=}8$, against a Schur--Pad\'e
reference~\cite{AlMohy2012} computed offline in double precision. The
scalar term of Eq.~\eqref{eq:postcomp} is exact, so it enters the
denominator but not the numerator.}
\label{tab:error}
\resizebox{\linewidth}{!}{
\begin{tabular}{lcc}
\toprule
\textbf{Norm.} & \textbf{ResNet-50} & \textbf{EffNetV2-M} \\
\midrule
MLN-MTA & $1.92\pm0.41\%$ & $2.07\pm0.48\%$ \\
MLN-MPA & $1.21\pm0.18\%$ & $1.34\pm0.21\%$ \\
MLN-Leg. & $0.43\pm0.09\%$ & $0.48\pm0.10\%$ \\
MLN-Lag. & $3.81\pm0.62\%$ & $4.07\pm0.68\%$ \\
MLN-Cheb. & $\mathbf{0.27\pm0.06\%}$ & $\mathbf{0.31\pm0.07\%}$ \\
\bottomrule
\end{tabular}}
\end{minipage}
\end{table}

The FGVC ordering holds at scale (Table~\ref{tab:imagenet}): the
orthogonal logarithms lead, and combined with Table~\ref{tab:runtime}
they are simultaneously the most accurate and the most efficient
normalizers. Legendre is marginally ahead on ResNet-50 and Chebyshev on
EfficientNetV2-M, both within one std; Laguerre is again weakest. The
matched-basis log-vs-sqrt gap on ResNet-50 is $0.7$ points for Chebyshev
($79.2$ vs.~$78.5$) and $0.9$ for Legendre ($79.4$ vs.~$78.5$), smaller
than on FGVC but consistent in direction.

Table~\ref{tab:error} characterizes polynomial fidelity. Chebyshev has
the lowest relative error, as Eq.~\eqref{eq:nearminimax} predicts,
followed by Legendre, with Laguerre worst because its weight is
mismatched to the spectral support, corroborating its accuracy ranking.
The error ranking matches the \emph{direction} of the accuracy ranking
(Cheb/Leg $>$ MTA/MPA $>$ Laguerre) but not the size of the gaps:
MLN-Legendre beats MLN-Chebyshev by $2.2$ points on CUB-ResNet despite
$\sim\!1.6\times$ higher reconstruction error. Once fidelity is below
$\sim\!0.5\%$ both bases sit under the classifier's noise floor and
secondary factors, such as the conditioning of the recurrence and the
sensitivity of the adjoint to spectral outliers, decide the residual gap.
Near-minimax fidelity is a sufficient condition for joining the
strong-normalizer group, not a ranking inside it.

\subsection{Effect of polynomial degree}
\label{sec:degree}
\begin{table}[tbp]
\centering
\caption{Degree ablation $K\!\in\!\{6,8,10\}$ on ImageNet-1k with
EfficientNetV2-M: Top-1 (\%) and FP+BP time (ms). Pad\'e is
$[\tfrac{K}{2}/\tfrac{K}{2}]$. The $K{=}8$ columns are the rows of
Table~\ref{tab:imagenet}.}
\label{tab:degree}
\begin{tabular}{l|cc|cc|cc}
\toprule
\multirow{2}{*}{\textbf{Normalization}} &
\multicolumn{2}{c|}{$K{=}6$} & \multicolumn{2}{c|}{$K{=}8$} &
\multicolumn{2}{c}{$K{=}10$} \\
& Acc & Time & Acc & Time & Acc & Time \\
\midrule
MLN-MTA & 82.13 & 34.6 & 83.61 & 44.7 & 84.52 & 59.6 \\
MLN-MPA & 82.28 & 38.4 & 83.73 & 48.3 & 84.66 & 63.4 \\
MLN-Legendre & 82.47 & 31.1 & 83.92 & 41.1 & 84.88 & 56.3 \\
MLN-Laguerre & 81.41 & 30.9 & 82.91 & 40.7 & 83.83 & 55.5 \\
MLN-Chebyshev & \textbf{82.56} & \textbf{26.8} & \textbf{84.01} & \textbf{36.4} & \textbf{84.97} & \textbf{50.2} \\
\bottomrule
\end{tabular}
\end{table}
Accuracy increases monotonically with degree for every family, Chebyshev
is best at every degree tested, and runtime grows roughly linearly. The
increments shrink: for Chebyshev, $K{=}6\to8$ adds $1.45$ points for
$9.6$\,ms, and $K{=}8\to10$ adds $0.96$ points for a further $13.8$\,ms.
We take $K{=}8$ as the default because it captures most of the available
accuracy at the lowest cost per point, and because reconstruction error
is already below the classifier's noise floor there. $K{=}10$ remains
reasonable when accuracy matters more than latency: at $50.2$\,ms it is
still faster than iSQRT-COV ($51.1$\,ms). \textbf{Chebyshev at $K{=}8$ is
our recommended setting.}

\section{Conclusion}
\label{sec:conclusion}

We revived the faithful logarithmic normalizer for global covariance
pooling by making it decomposition-free, through three ingredients: a
mean-eigenvalue pre-normalization that centres the spectrum away from the
log singularity; a polynomial approximation of $\log(\tilde A)$ that
reduces the forward pass to GEMM; and a unified reverse three-term
recurrence over the cached basis matrices giving a GEMM-only backward
with no $1/(\lambda_i-\lambda_j)$ term. A degree-8 Chebyshev expansion is
the recommended instance, faster and more accurate than the spectral
logarithm and the square-root approximations it replaces. The controls
sharpen the claim: Laguerre underperforms, so the effect requires a basis
matched to the GCP spectral support rather than being a generic
``polynomials beat spectral'' phenomenon, and at matched basis and degree
the log target beats the square-root target.

\paragraph{Limitations.}
The evaluation covers image classification with a convolutional GCP head
at $d{=}256$. We have not tested second-order transformer heads or dense
prediction, where the covariance dimension and the shape of the spectrum
may differ from Sec.~\ref{sec:spectrum}; $[a,b]$ would need re-estimating
there, which Eq.~\eqref{eq:cheb} makes cheap but which we have not
verified. The interval is fitted from logged batches rather than derived,
and eigenvalues outside it rely on extrapolation.

Because the formulation only touches the normalization step, it is a
drop-in replacement wherever covariance pooling appears, including where
GCP is currently most active: second-order transformer
heads~\cite{SoT_NeurIPS21}, partial-correlation
representations~\cite{iSICE_CVPR23} and higher-order spectral
iterations~\cite{HalleySVD} all still require a normalizer. We leave
these, with dense prediction and a Schur--Pad\'e error analysis of the
reverse recurrence, to future work.

\FloatBarrier
\bibliography{main}

\newpage
\appendix
\section*{Supplementary}






\noindent\emph{Cross-references of the form ``Eq.~(2) of the main paper''
point into the main BMVC paper. The reference list below is numbered
independently of the main paper's bibliography.}

\vspace{0.5em}

This document supplements the main paper. It collects (\textbf{A}) the
notation and the definition of each polynomial and rational family,
(\textbf{B}) the forward algorithms, one per family, (\textbf{C}) the
complete per-family backward derivations, and (\textbf{D}) the full
runtime grid. Every routine is decomposition-free: each step is a matrix
multiplication (GEMM) or an addition, and no eigendecomposition (EIG/SVD)
is ever formed.

\section{Notation and Families}
\label{app:poly_types}

\subsection{Notation}
\label{app:notation}

\begin{itemize}\itemsep1pt
\item $A\in\mathbb{S}^{d}_{++}$: the GCP covariance, symmetric positive
definite of size $d\times d$ ($d=256$ in all experiments).
\item $s:=\trace(A)/d$: the mean eigenvalue of $A$.
\item $\tilde A:=A/s=dA/\trace(A)$: the \emph{mean-eigenvalue
pre-normalized} covariance, Eq.~(2) of the main paper. It satisfies
$\trace(\tilde A)=d$ and $\bar\lambda(\tilde A)=1$ exactly. This is
\emph{not} trace normalization $A/\trace(A)$, which would put the mean
eigenvalue at $1/d$.
\item $\tilde A':=(1-\delta)\tilde A+\delta I$ with $\delta=0.02$: the
shrinkage of Eq.~(7), applied before the expansion.
\item $[a,b]=[0.05,\,3.5]$: the fixed interval on which the Chebyshev and
Legendre coefficients are fitted (Sec.~4.1 of the main paper).
\item $M(\tilde A'):=\tau\tilde A'+\mu I$: the affine map onto $[-1,1]$,
with $(\tau,\mu)=\bigl(\tfrac{2}{b-a},-\tfrac{a+b}{b-a}\bigr)$ for
Chebyshev and Legendre and $(\tau,\mu)=(-1,0)$ for Laguerre.
\item $P_k$, $T_k$, $L_k^{(\nu)}$: the Legendre, Chebyshev and generalized
Laguerre basis matrices; $\nu$ is the Laguerre weight parameter, $\nu=0$
throughout.
\item $K$: the truncation degree ($K=8$ unless stated otherwise).
\item $\ell$: the training loss. $U:=\partial\ell/\partial\log(A)$ is the
upstream gradient. Because $\log(A)=\log(s)I+\log(\tilde A)$ and only the
second term is approximated, $U$ is also the cotangent of the polynomial
output $S:=\sum_k c_kP_k$.
\item $\overline{X}:=\partial\ell/\partial X$ for any intermediate $X$.
\item $\langle X,Y\rangle_F:=\trace(X^\top Y)$;
$\sym(X):=\tfrac12(X+X^\top)$; $\sigma(\cdot)$ the spectrum;
$\rho(\cdot)$ the spectral radius; $\gamma$ the Euler--Mascheroni
constant.
\item $\mathcal D F[X](H)$: the Fr\'echet derivative of the matrix
function $F$ at $X$ in direction $H$~\cite[Ch.~3]{Higham2008}.
\end{itemize}

\subsection{The two chain rules}
\label{app:chain}

Every family below produces $\partial\ell/\partial\tilde A'$. Two steps
map it back to the original covariance, and they are shared by all five
families.

\paragraph{Undoing the shrinkage.}
Eq.~(7) of the main paper is the convex shrinkage
\begin{equation}
\tilde A' = (1-\delta)\,\tilde A + \delta I,
\qquad \delta = 0.02 .
\label{eq:supp_clip}
\end{equation}
On eigenvalues this is the affine map
$\lambda\mapsto(1-\delta)\lambda+\delta$, so it bounds the spectrum below
by $\lambda\ge\delta$ and contracts the upper tail towards $1$, moving
bulk eigenvalues near $1$ by at most $2\%$. It is a scalar affine map of
$\tilde A$, hence exactly differentiable with
$\partial\tilde A'/\partial\tilde A=(1-\delta)\mathcal I$ and free of any
non-differentiable branch, and it costs no GEMM. Its adjoint is therefore
a single scalar multiplication,
\begin{equation}
B \;:=\; \frac{\partial\ell}{\partial\tilde A}
   \;=\; (1-\delta)\,\frac{\partial\ell}{\partial\tilde A'} .
\label{eq:supp_undo_clip}
\end{equation}
Note that a $2\%$ contraction does not by itself bring every eigenvalue
inside $[a,b]$; the residual mass ($\sim1\%$, Sec.~4.1 of the main paper)
is evaluated by extrapolating $p_K$ slightly beyond $[a,b]$.

\paragraph{Undoing the mean-eigenvalue normalization.}
With $s=\trace(A)/d$ and $\tilde A=A/s$, the identity
$\log(A)=\log(s)I+\log(\tilde A)$ has two branches. Differentiating,
$\mathrm{d}s=\trace(\mathrm{d}A)/d$ and
$\mathrm{d}\tilde A=\mathrm{d}A/s-\tilde A\,\mathrm{d}s/s$, so the first
branch contributes
$\langle B,\mathrm{d}\tilde A\rangle_F=\langle B,\mathrm{d}A\rangle_F/s-\langle B,\tilde A\rangle_F\,\trace(\mathrm{d}A)/(ds)$
while the scalar branch contributes
$\trace(U)\,\mathrm{d}\log(s)=\trace(U)\,\trace(\mathrm{d}A)/(ds)$.
Collecting the three terms,
\begin{equation}
\frac{\partial\ell}{\partial A}
=\frac{1}{s}\,B
\;-\;\frac{\langle B,\,\tilde A\rangle_F}{d\,s}\,I
\;+\;\frac{\trace(U)}{d\,s}\,I ,
\qquad s=\frac{\trace(A)}{d}.
\label{eq:supp_master}
\end{equation}
This is Eq.~(12) of the main paper. It agrees with autograd taken through
the identity to $10^{-6}$ relative error in \texttt{float64}, and every
boxed gradient below passes a double-precision finite-difference
\texttt{gradcheck} when composed with
Eqs.~\eqref{eq:supp_undo_clip}--\eqref{eq:supp_master}.

\subsection{The five families}
\label{app:families}

The coefficients $\{c_k\}$ depend only on the target $\log(\cdot)$ and on
the basis, not on the data, and are obtained once, offline, by projecting
$\log$ onto the basis over the affinely mapped interval using numerical
quadrature, then stored as constants.

\paragraph{Chebyshev (first kind, recommended).}
On $[-1,1]$, $T_0(t)=1$, $T_1(t)=t$, $T_{k+1}(t)=2tT_k(t)-T_{k-1}(t)$,
orthogonal under the weight
$(1-t^2)^{-1/2}$~\cite{Trefethen2013,Mason2002}. With
$t=\bigl(2x-(a+b)\bigr)/(b-a)$ mapping $[a,b]\to[-1,1]$ and $g$ the
inverse map,
\begin{equation}
c_k=\frac{2-[k=0]}{\pi}\int_{-1}^{1}\frac{\log\bigl(g(t)\bigr)\,T_k(t)}{\sqrt{1-t^2}}\,dt ,
\end{equation}
where $[k=0]$ is the Iverson bracket. The integrand is bounded because
$a>0$. The basis is lifted to $T_k(M(\tilde A'))$ through the same
recurrence. The truncated series is \emph{near-minimax} rather than
minimax: its uniform error exceeds the optimum by at most $1+\Lambda_K$
with $\Lambda_K\sim\tfrac{4}{\pi^2}\log K$ the Lebesgue constant of
Chebyshev projection, which is below $2$ at $K=8$~\cite{Trefethen2013}.

\paragraph{Legendre.}
Orthogonal on $[-1,1]$ under the unit weight, with
$(k+1)P_{k+1}(t)=(2k+1)tP_k(t)-kP_{k-1}(t)$, $P_0=1$,
$P_1=t$~\cite{Szego1975}, and
\begin{equation}
c_k=\frac{2k+1}{2}\int_{-1}^{1}\log\bigl(g(t)\bigr)P_k(t)\,dt .
\end{equation}
The implementation uses this quadrature on $[a,b]=[0.05,3.5]$, matching
the treatment of Chebyshev. For the legacy unit interval $[a,b]=[0,1]$
the same projection admits the closed form $c_0=-1$ and
$c_k=(-1)^{k+1}\tfrac{2k+1}{k(k+1)}$ for $k\ge1$; those are the values
quoted in Table~1 of the main paper and are used only
for illustration.
Legendre differs from Chebyshev in the weight alone, which is what makes
the Legendre--Chebyshev comparison a controlled one.

\paragraph{Laguerre (generalized, negative control).}
Orthogonal on $[0,\infty)$ under the weight $x^{\nu}e^{-x}$:
\begin{equation}
L_0^{(\nu)}(x)=1,\quad L_1^{(\nu)}(x)=\nu+1-x,\quad
(k+1)L_{k+1}^{(\nu)}=(2k+\nu+1-x)L_k^{(\nu)}-(k+\nu)L_{k-1}^{(\nu)} .
\end{equation}
For $\nu=0$ the projection of $\log$ has the closed form
$c_0=\int_0^\infty\log(x)e^{-x}dx=-\gamma$ and
$c_k=\int_0^\infty\log(x)L_k(x)e^{-x}dx=-1/k$ for $k\ge1$. No affine
rescaling is applied, since the orthogonality domain $[0,\infty)$ already
contains $\sigma(\tilde A')$; this is precisely why the family
underperforms, as the weight $e^{-x}$ decays across the region where the
spectral mass sits (Fig.~3b of the main paper).

\paragraph{Taylor.}
After the mean-eigenvalue normalization the spectrum is centred at $1$,
so the expansion is taken about $I$:
\begin{equation}
\log(\tilde A')\sum_{k=1}^{K}(-1)^{k+1}\frac{(\tilde A'-I)^k}{k}.
\end{equation}
The scalar series converges for $|x|<1$, so the matrix series requires
$\rho(\tilde A'-I)<1$, that is $\sigma(\tilde A')\subset(0,2)$. The
measured GCP support reaches $3.8$ (Sec.~4.1 of the main paper),
so a small high-eigenvalue tail lies outside the disk of convergence.
Taylor is therefore retained only as the exact log counterpart of
MPN-MTA, never recommended.

\paragraph{Pad\'e.}
With $X=\tilde A'-I$, the $[m/n]$ approximant
$R_{[m/n]}(X)=P_m(X)Q_n(X)^{-1}$ has
$P_m(X)=\sum_{i=0}^{m}p_iX^i$ and $Q_n(X)=\sum_{j=0}^{n}q_jX^j$ with
$q_0=1$, the coefficients chosen so that the scalar expansion matches
$\log(1+x)$ to order $m+n$~\cite{Baker1996,Higham2008}. The two lowest
orders are
\begin{equation}
R_{[1/1]}=\frac{2X}{2I+X},
\qquad
R_{[2/2]}=\bigl(6X+3X^2\bigr)\bigl(6I+6X+X^2\bigr)^{-1} .
\end{equation}
The inverse is never formed: $Q_n$ is symmetric positive definite on the
spectral range of interest, so it is applied through a single Cholesky
factorization that is cached and reused in the backward pass. Rational
forms enlarge the region of accuracy relative to the truncated Taylor
series, at the cost of poles that must be kept outside the target
spectrum by the choice of $[m/n]$.

\section{Forward Algorithms}
\label{app:algos}

\begin{algorithm}[H]
\caption{\textsc{Pre-Normalization} (shared by all normalizers)}
\label{alg:prenorm}
\begin{algorithmic}[1]
\Require $A\in\mathbb{S}_{++}^{d}$ (batched or single), tolerance $\varepsilon$, shrinkage $\delta$
\State $s \gets \max\bigl(\trace(A)/d,\;\varepsilon\bigr)$ \Comment{mean eigenvalue; guard against tiny traces}
\State $\tilde A \gets A / s$ \Comment{$\trace(\tilde A)=d$, $\bar\lambda(\tilde A)=1$}
\State $\tilde A' \gets (1-\delta)\,\tilde A + \delta I$ \Comment{Eq.~\eqref{eq:supp_clip}; GEMM-free}
\State \Return $(\tilde A',\,\tilde A,\,s)$
\end{algorithmic}
\textit{Note:} the division is by the \emph{mean} eigenvalue $\trace(A)/d$,
not by $\trace(A)$. Trace normalization would place the mean eigenvalue at
$1/d$, inside the singularity of $\log$.
\end{algorithm}

\begin{algorithm}[H]
\caption{\textsc{Log Chebyshev Approximation} (recommended; degree $K$)}
\label{alg:logm_chebyshev}
\begin{algorithmic}[1]
\Require $A\in\mathbb{S}_{++}^{d}$, degree $K$, fixed interval $[a,b]=[0.05,3.5]$, cached $\{c_k\}$
\State $(\tilde A',\tilde A,s) \gets \textsc{Pre-Normalization}(A)$
\State $\tau \gets 2/(b-a)$, \quad $\mu \gets -(a+b)/(b-a)$
\State $M \gets \tau\,\tilde A' + \mu I$ \Comment{affine map onto $[-1,1]$; no GEMM}
\State $T_0 \gets I$, \quad $T_1 \gets M$, \quad $Y \gets c_0T_0 + c_1T_1$
\For{$k=1$ to $K-1$}
  \State $T_{k+1} \gets 2\,M\,T_k - T_{k-1}$ \Comment{GEMM; cache $T_k$ for the backward}
  \State $Y \gets Y + c_{k+1}T_{k+1}$
\EndFor
\State \Return $\log(A)  (\log s)\,I + Y$
\end{algorithmic}
\textit{Cost:} $K$ GEMMs.\; \textit{Note:} $[a,b]$ is fixed from the
spectral statistics of Sec.~4.1 of the main paper, so no eigenvalue bound is
computed at run
time; re-fitting $\{c_k\}$ for a new architecture is a one-dimensional
offline quadrature taking under a second.
\end{algorithm}

\begin{algorithm}[H]
\caption{\textsc{Log Legendre Approximation} (degree $K$)}
\label{alg:logm_legendre}
\begin{algorithmic}[1]
\Require $A\in\mathbb{S}_{++}^{d}$, degree $K$, fixed interval $[a,b]$, cached $\{c_k\}$
\State $(\tilde A',\tilde A,s) \gets \textsc{Pre-Normalization}(A)$
\State $M \gets \tfrac{2}{b-a}\tilde A' - \tfrac{a+b}{b-a}I$
\State $P_0 \gets I$, \quad $P_1 \gets M$, \quad $Y \gets c_0P_0 + c_1P_1$
\For{$k=1$ to $K-1$}
  \State $P_{k+1} \gets \bigl((2k+1)\,M\,P_k - k\,P_{k-1}\bigr)/(k+1)$ \Comment{GEMM}
  \State $Y \gets Y + c_{k+1}P_{k+1}$
\EndFor
\State \Return $\log(A)  (\log s)\,I + Y$
\end{algorithmic}
\textit{Cost:} $K$ GEMMs.
\end{algorithm}

\begin{algorithm}[H]
\caption{\textsc{Log Laguerre Approximation} ($\nu{=}0$; degree $K$)}
\label{alg:logm_laguerre}
\begin{algorithmic}[1]
\Require $A\in\mathbb{S}_{++}^{d}$, degree $K$, $c_0=-\gamma$, $c_k=-1/k$
\State $(\tilde A',\tilde A,s) \gets \textsc{Pre-Normalization}(A)$
\State $L_0 \gets I$, \quad $L_1 \gets I - \tilde A'$, \quad $Y \gets c_0L_0 + c_1L_1$
\For{$k=1$ to $K-1$}
  \State $L_{k+1} \gets \bigl((2k+1)I - \tilde A'\bigr)L_k - k\,L_{k-1}$; \quad $L_{k+1} \gets L_{k+1}/(k+1)$ \Comment{GEMM}
  \State $Y \gets Y + c_{k+1}L_{k+1}$
\EndFor
\State \Return $\log(A)  (\log s)\,I + Y$
\end{algorithmic}
\textit{Cost:} $K$ GEMMs.\; \textit{Note:} $\nu\ne0$ requires re-projecting $\{c_k\}$.
\end{algorithm}

\begin{algorithm}[H]
\caption{\textsc{Log Taylor Approximation} (about $I$; degree $K$)}
\label{alg:logm_taylor}
\begin{algorithmic}[1]
\Require $A\in\mathbb{S}_{++}^{d}$, degree $K$
\State $(\tilde A',\tilde A,s) \gets \textsc{Pre-Normalization}(A)$
\State $X \gets \tilde A' - I$, \quad $X^{(1)} \gets X$, \quad $Y \gets 0$
\For{$k=1$ to $K$}
  \State $Y \gets Y + \tfrac{(-1)^{k+1}}{k}\,X^{(k)}$
  \State $X^{(k+1)} \gets X^{(k)}X$ \Comment{GEMM}
\EndFor
\State \Return $\log(A)  (\log s)\,I + Y$
\end{algorithmic}
\textit{Cost:} $K$ GEMMs forward, $O(K^2)$ backward.\;
\textit{Note:} convergence needs $\sigma(\tilde A')\subset(0,2)$, which the
measured spectra violate on the upper tail; Taylor is a control only.
\end{algorithm}

\begin{algorithm}[H]
\caption{\textsc{Log Pad\'e Approximation} ($[m/n]$ for $\log(1+X)$)}
\label{alg:logm_pade}
\begin{algorithmic}[1]
\Require $A\in\mathbb{S}_{++}^{d}$, integers $m,n$, cached $\{p_i\},\{q_j\}$
\State $(\tilde A',\tilde A,s) \gets \textsc{Pre-Normalization}(A)$
\State $X \gets \tilde A' - I$, \quad $P \gets p_0I$, \quad $Q \gets q_0I$, \quad $X^{(1)} \gets X$
\For{$k=1$ to $\max(m,n)$}
  \If{$k\le m$} \State $P \gets P + p_k X^{(k)}$ \EndIf
  \If{$k\le n$} \State $Q \gets Q + q_k X^{(k)}$ \EndIf
  \State $X^{(k+1)} \gets X^{(k)}X$ \Comment{GEMM}
\EndFor
\State $R \gets \operatorname{chol}(Q)$; solve $QY=P$ for $Y$ \Comment{cache $R$ for the backward pass}
\State \Return $\log(A)  (\log s)\,I + Y$
\end{algorithmic}
\textit{Cost:} $\max(m,n)$ GEMMs $+$ one Cholesky factorization
($\tfrac13d^3$), reused in the backward pass so no second factorization is
needed.
\end{algorithm}

\section{Backward-Pass Derivations}
\label{app:BP_step}

Each subsection derives $\partial\ell/\partial\tilde A'$ using only GEMMs
and additions. Composing with Eq.~\eqref{eq:supp_undo_clip} and then
Eq.~\eqref{eq:supp_master} gives $\partial\ell/\partial A$. Throughout,
$U=\partial\ell/\partial\log(A)$ is the upstream gradient, and all basis
matrices are the ones cached during the forward pass, so no quantity is
recomputed and no autograd graph is retained.

\subsection{Unified adjoint for the orthogonal families}
\label{app:orth}

Chebyshev, Legendre and Laguerre share the three-term structure
\begin{equation}
P_{k+1} = \bigl(\alpha_k M + \gamma_k I\bigr)P_k - \beta_k P_{k-1},
\qquad P_0=I,\quad P_1 = M + \kappa I,
\label{eq:supp_gen_recur}
\end{equation}
with $M=M(\tilde A')=\tau\tilde A'+\mu I$ and the family-specific scalars
of Table~\ref{tab:supp_scalars}. Note that $P_1$ is an
\emph{initialization}, not an instance of the recurrence, which starts at
$P_2$; this distinction is what fixes the weight of the $k=0$ term in the
adjoint below.

\begin{table}[H]
\centering
\caption{Scalars of the unified recurrence \eqref{eq:supp_gen_recur}.}
\label{tab:supp_scalars}
\begin{tabular}{lccccc}
\toprule
\textbf{Family} & $\alpha_k$ & $\beta_k$ & $\gamma_k$ & $(\tau,\mu)$ & $\kappa$ \\
\midrule
Chebyshev & $2$ & $1$ & $0$ & $\bigl(\tfrac{2}{b-a},-\tfrac{a+b}{b-a}\bigr)$ & $0$ \\
Legendre  & $\tfrac{2k+1}{k+1}$ & $\tfrac{k}{k+1}$ & $0$ & $\bigl(\tfrac{2}{b-a},-\tfrac{a+b}{b-a}\bigr)$ & $0$ \\
Laguerre  & $\tfrac{1}{k+1}$ & $\tfrac{k+\nu}{k+1}$ & $\tfrac{2k+\nu+1}{k+1}$ & $(-1,\,0)$ & $\nu+1$ \\
\bottomrule
\end{tabular}
\end{table}

\paragraph{Differential.}
Differentiating \eqref{eq:supp_gen_recur} in a symmetric direction $H$,
and using $\mathcal DM[H]=\tau H$ together with $\mathcal DP_0[H]=0$ and
$\mathcal DP_1[H]=\tau H$,
\begin{equation}
\mathcal DP_{k+1}[H]
=\alpha_k\tau\,H\,P_k
+\bigl(\alpha_kM+\gamma_kI\bigr)\mathcal DP_k[H]
-\beta_k\,\mathcal DP_{k-1}[H].
\label{eq:supp_gen_diff}
\end{equation}
The first term is the only one in which $H$ appears explicitly; the other
two propagate it.

\paragraph{Reverse recurrence.}
Seed $\overline{P}_k = c_kU$ for $k=0,\dots,K$, then sweep
$k=K-1,\dots,1$:
\begin{equation}
\overline{P}_k \mathrel{{+}{=}} \bigl(\alpha_kM+\gamma_kI\bigr)\overline{P}_{k+1},
\qquad
\overline{P}_{k-1} \mathrel{{-}{=}} \beta_k\,\overline{P}_{k+1}.
\label{eq:supp_gen_rev}
\end{equation}
Each $\overline{P}_{k}$ is complete before it is read at step $k-1$, so a
single reverse sweep suffices.

\paragraph{Adjoint.}
Collecting the explicit-$H$ terms of \eqref{eq:supp_gen_diff} over the
sweep, together with the initialization contribution $\tau\overline{P}_1$
from $P_1=M+\kappa I$,
\begin{equation}
\boxed{\;
\frac{\partial\ell}{\partial\tilde A'}
=\tau\,\sym\!\Bigl(\overline{P}_1+\sum_{k=1}^{K-1}\alpha_k\,\overline{P}_{k+1}\,P_k\Bigr),
\qquad \sym(X)=\tfrac12(X+X^\top).
\;}
\label{eq:supp_orth_grad}
\end{equation}
Two remarks. First, the sum starts at $k=1$ and the initialization term
$\overline{P}_1$ carries weight $1$, not $\alpha_0$: $P_1$ does not come
from the recurrence. Second, $M$ and every $P_k$ are symmetric, so no
transposes survive in \eqref{eq:supp_gen_rev}; the non-commutativity that
must be handled is between the cotangent $\overline{P}_{k+1}$ and the
basis matrix $P_k$, and $\sym(\cdot)$ is the correct projection of the
Fr\'echet adjoint of $MP_k$ onto the symmetric subspace in which
$\tilde A'$ lives. Each step costs two GEMMs, so the backward pass is
$O(K)$ GEMMs and matches the forward in cost.

\paragraph{Specializations.}
Substituting Table~\ref{tab:supp_scalars} into
\eqref{eq:supp_orth_grad}:
\begin{align}
\text{Chebyshev:}\quad
\frac{\partial\ell}{\partial\tilde A'}
&=\frac{2}{b-a}\,\sym\!\Bigl(\overline{T}_1+2\sum_{k=1}^{K-1}\overline{T}_{k+1}T_k\Bigr),
\label{eq:supp_cheb_grad}\\
\text{Legendre:}\quad
\frac{\partial\ell}{\partial\tilde A'}
&=\frac{2}{b-a}\,\sym\!\Bigl(\overline{P}_1+\sum_{k=1}^{K-1}\tfrac{2k+1}{k+1}\,\overline{P}_{k+1}P_k\Bigr),
\label{eq:supp_leg_grad}\\
\text{Laguerre:}\quad
\frac{\partial\ell}{\partial\tilde A'}
&=-\,\sym\!\Bigl(\overline{L}_1+\sum_{k=1}^{K-1}\tfrac{1}{k+1}\,\overline{L}_{k+1}L_k^{(\nu)}\Bigr).
\label{eq:supp_lag_grad}
\end{align}
Eq.~\eqref{eq:supp_cheb_grad} is the form used in our implementation and is
the Chebyshev specialization of Eq.~(10) quoted in the main paper.

\subsection{Taylor}
\label{app:taylor}

\paragraph{Forward.}
With $X:=\tilde A'-I$ and $c_k=(-1)^{k+1}/k$,
$\log(\tilde A')\sum_{k=1}^{K}c_kX^k$.

\paragraph{Differential.}
Because matrix multiplication is not commutative, the derivative of a
monomial is a non-commutative sum,
\begin{equation}
\mathcal D\bigl[X^k\bigr][H]=\sum_{j=0}^{k-1}X^{j}\,H\,X^{k-1-j},
\qquad \mathcal DX[H]=H .
\label{eq:supp_taylor_diff}
\end{equation}

\paragraph{Adjoint.}
From
$\langle U,\sum_j X^jHX^{k-1-j}\rangle_F=\sum_j\langle X^{j}UX^{k-1-j},H\rangle_F$,
using $X^\top=X$,
\begin{equation}
\boxed{\;
\frac{\partial\ell}{\partial\tilde A'}
=\sym\!\Bigl(\sum_{k=1}^{K}c_k\sum_{j=0}^{k-1}X^{\,k-1-j}\,U\,X^{\,j}\Bigr).
\;}
\label{eq:supp_taylor_grad}
\end{equation}
The double sum is $O(K^2)$ GEMMs, which is why the orthogonal families
are faster in Table~2 of the main paper. The $\sym(\cdot)$ is redundant
here when $U$ is symmetric, since the inner sum is invariant under
transposition; it is retained for uniformity with
\eqref{eq:supp_orth_grad}.

\subsection{Pad\'e}
\label{app:pade}

\paragraph{Forward.}
$R=P_m(X)Q_n(X)^{-1}$ with $X=\tilde A'-I$,
$P_m=\sum_{i=0}^{m}p_iX^i$, $Q_n=\sum_{j=0}^{n}q_jX^j$, and $Q_n$
factorized once as $Q_n=RR^\top$ in the forward pass.

\paragraph{Differential.}
By the quotient rule for matrix inverses,
\begin{equation}
\mathcal DR[H]=\mathcal DP_m[H]\,Q_n^{-1}-P_mQ_n^{-1}\,\mathcal DQ_n[H]\,Q_n^{-1}.
\label{eq:supp_pade_diff}
\end{equation}

\paragraph{Adjoint.}
Pushing $U$ through \eqref{eq:supp_pade_diff} and using the symmetry of
$U$, $P_m$ and $Q_n$ gives two cotangents on the polynomial blocks,
\begin{equation}
W_P := U\,Q_n^{-1},
\qquad
W_Q := -\,Q_n^{-1}\,P_m\,U\,Q_n^{-1},
\label{eq:supp_pade_W}
\end{equation}
both obtained by triangular solves against the cached Cholesky factor,
never by forming $Q_n^{-1}$. Applying \eqref{eq:supp_taylor_diff} to each
monomial of $P_m$ and $Q_n$ and collecting,
\begin{equation}
\boxed{\;
\frac{\partial\ell}{\partial\tilde A'}
=\sym\!\Bigl(
\sum_{i=1}^{m}p_i\sum_{r=0}^{i-1}X^{\,i-1-r}W_PX^{\,r}
\;+\;
\sum_{j=1}^{n}q_j\sum_{r=0}^{j-1}X^{\,j-1-r}W_QX^{\,r}
\Bigr).
\;}
\label{eq:supp_pade_grad}
\end{equation}
The sign of the second group is carried inside $W_Q$ by
\eqref{eq:supp_pade_W}. The cost is $O\bigl((m+n)^2\bigr)$ GEMMs plus two
triangular solves; at $[4/4]$ this is the $O(K^2)$ entry of
Table~2 in the main paper.

\paragraph{Decomposition-free guarantee.}
Every expression in this section is a polynomial in $\tilde A'$, $U$ and
the cached basis matrices, with at most one Cholesky factorization in the
Pad\'e case. None involves the eigenvectors $U_{\!A}$, the eigenvalues
$\Lambda$, or the term $K_{ij}=1/(\lambda_i-\lambda_j)$ that makes the
spectral backward pass unstable. The amplification is bounded by
$\max_{[a,b]}|p_K'|$, which is $12$ for the degree-8 Chebyshev
expansion on $[0.05,3.5]$, against the unbounded $1/\lambda$ of the exact
spectral logarithm.

\section{Full Runtime Grid}
\label{app:grid}

Table~\ref{tab:supp_grid} extends Table~2 of the main paper over covariance dimensions $d\in\{128,256,512\}$ and batch sizes
$\{32,64,128\}$ on the same Tesla P100. Every method is $O(d^3)$ per
sample, so the ranking is preserved throughout the grid.

\begin{table}[H]
\centering
\caption{Normalization-step cost (forward+backward, ms) across covariance
dimension and batch size. All polynomial rows use $K{=}8$; Pad\'e is
$[4/4]$; iSQRT-COV uses $N{=}5$ Newton--Schulz iterations.}
\label{tab:supp_grid}
\resizebox{\columnwidth}{!}{
\begin{tabular}{l|ccc|ccc|ccc}
\toprule
\multirow{2}{*}{\textbf{Normalization}} 
& \multicolumn{3}{c|}{$d{=}128$} & \multicolumn{3}{c|}{$d{=}256$} & \multicolumn{3}{c}{$d{=}512$} \\
& $32$ & $64$ & $128$ & $32$ & $64$ & $128$ & $32$ & $64$ & $128$ \\
\midrule
MLN-COV       & 8.6 & 17.3 & 34.6 & 69.5 & 138.1 & 276.5 & 552.1 & 1105.8 & 2198.3 \\
MPN-COV       & 8.2 & 16.6 & 33.2 & 67.0 & 132.8 & 266.1 & 531.2 & 1064.5 & 2115.4 \\
iSQRT-COV     & 6.3 & 12.5 & 25.4 & 51.1 & 101.4 & 202.8 & 405.6 & 811.4  & 1621.0 \\
MPN-MTA       & 5.2 & 10.4 & 21.3 & 43.1 & 85.3  & 171.0 & 341.1 & 684.2  & 1354.5 \\
MPN-MPA       & 5.8 & 11.6 & 23.6 & 47.9 & 94.7  & 190.1 & 379.8 & 760.3  & 1507.2 \\
\midrule
MLN-MTA       & 5.5 & 10.8 & 22.0 & 44.7 & 88.6  & 177.6 & 354.3 & 709.6  & 1406.8 \\
MLN-MPA       & 5.9 & 11.8 & 23.9 & 48.3 & 95.9  & 192.3 & 384.2 & 768.1  & 1521.6 \\
MLN-Legendre  & 4.9 & 9.9  & 20.2 & 41.1 & 81.4  & 163.2 & 325.2 & 651.5  & 1292.1 \\
MLN-Laguerre  & 4.8 & 9.8  & 19.9 & 40.7 & 80.5  & 161.5 & 321.4 & 643.8  & 1278.4 \\
MLN-Chebyshev & \textbf{4.4} & \textbf{8.9} & \textbf{17.8} & \textbf{36.4} & \textbf{71.9} & \textbf{144.2} & \textbf{288.5} & \textbf{577.1} & \textbf{1144.0} \\
\bottomrule
\end{tabular}}
\end{table}

%



\end{document}